\documentclass{article}

\PassOptionsToPackage{numbers,sort&compress}{natbib}
\usepackage[main,final]{neurips_2026}

\usepackage[utf8]{inputenc} 
\usepackage[T1]{fontenc}    
\usepackage{hyperref}       
\usepackage{url}            
\usepackage{booktabs}       
\usepackage{amsfonts}       
\usepackage{nicefrac}       
\usepackage{microtype}      
\usepackage{xcolor}         
\usepackage{amsmath}
\usepackage{amssymb}
\usepackage{graphicx}
\usepackage{multirow}
\usepackage{subcaption}
\usepackage{algorithm}
\usepackage{algorithmic}
\usepackage{xspace}

\newcommand{\ours}{TwinPrune\xspace}

\graphicspath{{figs/}}

\title{\ours{}: Density-Aware Two-Phase \\ Token Pruning for Vision-Language Models}

\author{
  Kai Liu$^{1}$\thanks{Equal Contribution},\enspace
  Anqi Li$^{1*}$,\enspace
  Junxian Li$^{1}$,\enspace \\
  \textbf{Zhixin Wang}$^{2}$,\enspace
  \textbf{Zhikai Chen}$^{2}$,\enspace
  \textbf{Renjing Pei}$^{2}$,\enspace
  \textbf{Yulun Zhang}$^{1}$\thanks{Corresponding author: Yulun Zhang, yulun100@gmail.com} \enspace
  \\
  \textsuperscript{1}Shanghai Jiao Tong University,\enspace
  \textsuperscript{2}Huawei Technologies Ltd
}

\begin{document}
\setlength{\abovedisplayskip}{2pt}
\setlength{\belowdisplayskip}{2pt}

\maketitle

\begin{figure}[H]
\centering
\includegraphics[width=\linewidth]{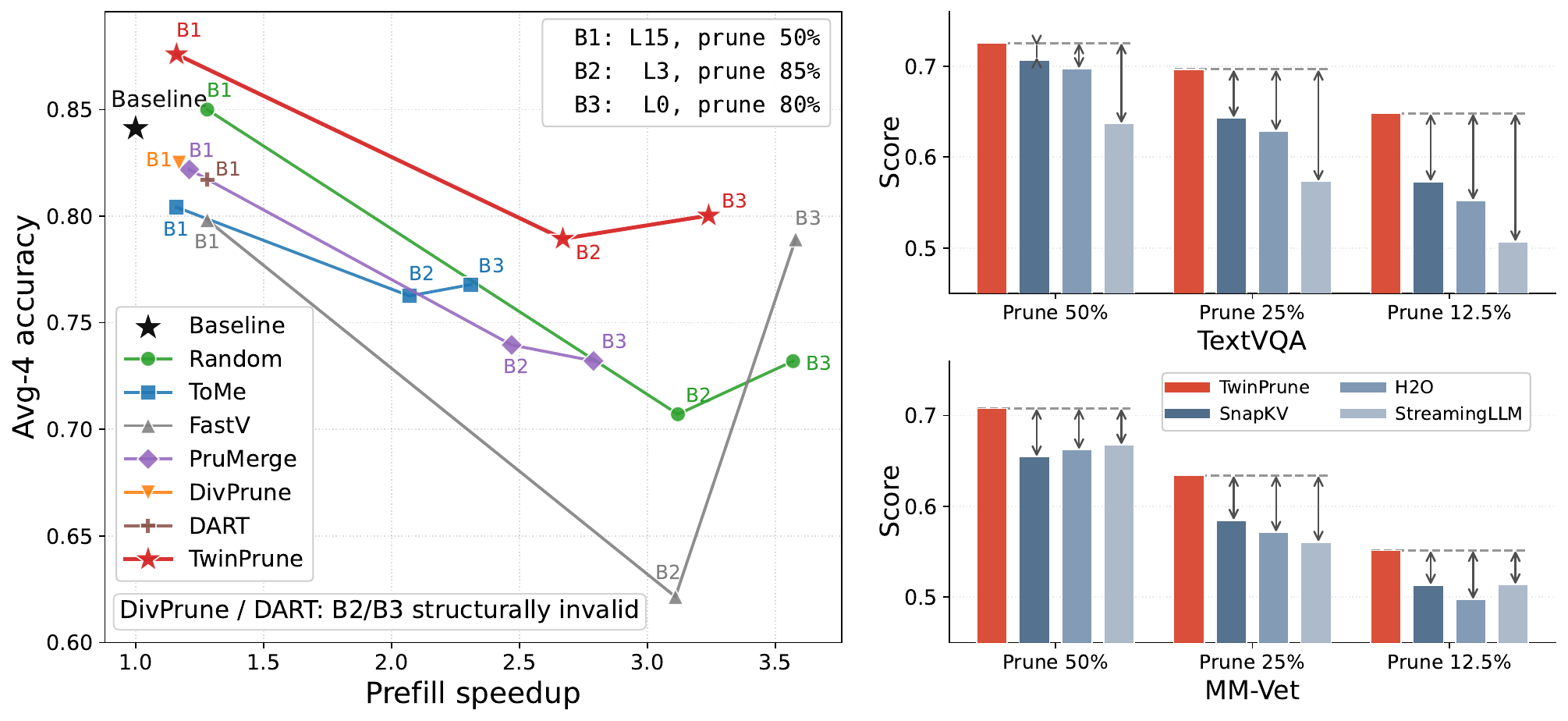}
\caption{\textbf{Two-Stage Pruning in TwinPrune.} TwinPrune improves VLM inference through a two-stage pruning framework using a single unified pruning signal. The left panel evaluates visual token pruning in the prefilling stage, showing better accuracy–speedup trade-offs, while the right panel evaluates KV cache pruning on long-output benchmarks in the decode stage. Both panels report results on InternVL-3.5-8B under the same evaluation setting.}
\label{fig:teaser}
\end{figure}

\begin{abstract}
We propose a unified spatial-density signal that scores token importance for \emph{both} phases of inference, instantiated as a bipartite merge during prefill and a KV eviction during decode, with measurable gains in each phase. To verify the two phases independently, we partition benchmarks into two diagnostic roles by their scoring rule. \emph{Prefill probes} such as MMBench, POPE, and ScienceQA score only the first decoded token, so their accuracy reflects only the argmax of prefill logits, computed before any decode-phase KV eviction takes effect. \emph{Full-pipeline probes} such as TextVQA, DocVQA, and MM-Vet score the full decoded sequence and therefore reveal both phases. We expose a fundamental evaluation flaw: published claims of decode-phase token pruning validated only on prefill probes have measured prefill quality, not decode quality. Empirically, removing all visual KV at decode leaves prefill-probe accuracy essentially unchanged while collapsing TextVQA and MM-Vet by tens of points, and the insensitivity is a property of the scoring rule, not of the compression strength. Across seven model variants from the InternVL-3.5, Qwen3-VL, and LLaVA-1.5 families and the three full-pipeline-probe benchmarks, we deliver an honest three-axis Pareto over compute, memory, and accuracy. Density-based prefill compression is competitive on the accuracy frontier, scales linearly in the visual-token count, and is roughly two orders of magnitude faster in token selection at high resolution. On the decode side, density top-$k$ KV eviction wins every TextVQA and MM-Vet cell against the strongest published baselines H2O, SnapKV, and StreamingLLM across three keep ratios. The advantage widens to $9.6$ percentage points over H2O at $12.5\%$ keep on TextVQA and holds $7$ to $10$ percentage points across the InternVL-3.5 series from 2B to 38B. Compression also scales gracefully with model size: every matched 8B-to-38B transition we measure reduces, rather than amplifies, the accuracy loss. We recommend MM-Vet or DocVQA as the minimum full-pipeline-probe test for any decode-phase claim.
Our code is available at \url{https://github.com/Kai-Liu001/TwinPrune}.
\end{abstract}

\section{Introduction}\label{sec:intro}

Vision-Language Models (VLMs) now drive a fast-growing set of applications such as document understanding, embodied agents, multi-step visual reasoning, and GUI control. Each of these demands processing hundreds to thousands of visual tokens through tens of transformer layers. As image resolution and tile count rise, two efficiency bottlenecks tighten in tandem. The first is \emph{prefill compute}, dominated by encoding all visual tokens through the language model. The second is \emph{decode memory}, dominated by the per-layer key/value cache that must persist for autoregressive generation. Whether a model can be deployed at all, whether on a memory-constrained device, in a low-latency interactive setting, or under a high-throughput serving regime, is increasingly determined less by how the model is trained and more by how aggressively redundant visual tokens can be removed at inference time.

The token-pruning literature has expanded rapidly in response. Prefill-time methods such as FastV~\cite{chen2024fastv}, ToMe~\cite{bolya2023tome}, DivPrune~\cite{divprune2025}, and DART~\cite{dart2025} target the compute axis, while decode-time methods such as H2O~\cite{h2o2023}, StreamingLLM~\cite{streamingllm}, and SnapKV~\cite{snapkv} target the memory axis. These works converge on a striking shared finding: aggressive compression preserves accuracy almost perfectly, often within a fraction of a point of the uncompressed baseline. The implication, repeated across dozens of papers, is that visual tokens are deeply redundant and can be cut without harm. If that picture were correct, token pruning would be a solved problem, a routine deployment knob rather than an open question.

We argue this consensus rests on a measurement artifact. The benchmarks most commonly used to validate VLM pruning are dominated by single-token outputs such as a multiple-choice letter or a yes/no answer. Their score depends only on the first decoded logit, which is fixed by the prefill pass before any decode-phase KV eviction is executed. By construction, decode-phase eviction therefore cannot move those scores, so what is measured under a ``decode-phase'' label is in fact prefill-phase quality. Whether a method preserves behavior under longer generation, in which information has to be retrieved from the cache during decode, is a separate question, and it is not the one those benchmarks answer. The same critique applies in a milder form to prefill compression: prefill-probe scores see only the argmax shift, not the distributional damage that compression does to longer continuations.

\begin{figure}[t]
\centering
\includegraphics[width=\linewidth]{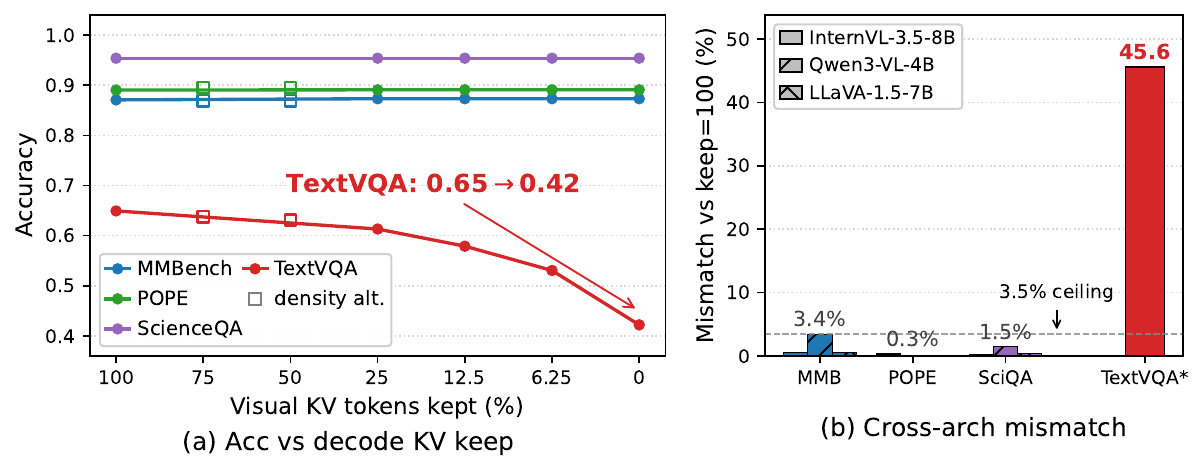}
\caption{Prefill-probe benchmarks are structurally insensitive to decode-phase KV eviction across three VLM families. Left: accuracy vs.\ visual KV kept on InternVL-3.5-8B. Right: prediction mismatch at \textsc{Dall} grouped by (benchmark, model). Details in Sec.~\ref{sec:methodology}.}
\label{fig:methodology}
\vspace{-2mm}
\end{figure}

We make the gap concrete by running a VLM with \emph{all} visual KV removed at decode time. On the prefill-probe benchmarks (MMBench, POPE, ScienceQA) the accuracy is nearly unchanged. On full-pipeline-probe benchmarks such as TextVQA and MM-Vet, accuracy drops sharply, and prediction strings flip on a large fraction of samples (Tab.~\ref{tab:mismatch}, Fig.~\ref{fig:methodology}, Sec.~\ref{sec:methodology}). The model \emph{is} changing its behavior under compression, but the prefill-probe benchmarks simply cannot register that change.

This re-scopes what an honest unified compression framework should look like and what it should be evaluated on. We propose a spatial-density signal $\rho$ that scores each visual token by the local non-redundancy of its $w{\times}w$ neighborhood on the regular spatial grid, computed once at an early decoder layer from token-to-token cosine distances. The same $\rho$ drives both compression phases. During prefill, we deploy it inside a density-tilted farthest-point recurrence that combines $k$-center coverage with density priority through a single tilt scalar $\beta$, bipartitions the visual tokens, and merges the discarded-token features into the kept set before passing the reduced sequence forward. During decode, the same $\rho$ values are reused as a top-$k$ score for visual-KV eviction, with no new attention pattern computed. The two phases can be applied independently or jointly, and we supply a three-axis (compute, memory, accuracy) Pareto study across seven model variants and three benchmarks.

Empirically, the unified density signal is competitive on the prefill accuracy frontier and is roughly two orders of magnitude faster in token selection at high resolution, and the implementation, with a Triton-fused farthest-point inner loop and a cached CUDA graph, delivers a $3.24\times$ end-to-end prefill speedup at the most aggressive operating point on InternVL-3.5-8B. On the decode side, density top-$k$ KV eviction wins every TextVQA and MM-Vet cell at three keep ratios against H2O, SnapKV, and StreamingLLM, with the gap widening to $9.6$ percentage points over H2O at $12.5\%$ keep on TextVQA and holding $7$ to $10$pp across the InternVL-3.5 series from 2B to 38B. Compression cost \emph{does not grow} with model size in the 8B-to-38B regime.
Our contributions are as follows:

\begin{enumerate}
\item \textbf{Methodology finding (Sec.~\ref{sec:methodology}):} prefill-probe benchmarks (those scoring only the first decoded token) structurally cannot measure decode-phase compression. We provide the proof, an extended cross-architecture mismatch table, and a recommendation for how the field should evaluate decode-phase claims going forward.
\item \textbf{Unified density framework (Sec.~\ref{sec:framework}):} a single spatial-density signal that scores token importance for both compression phases, with one kernel and two deployment points, guiding bipartite merge during prefill and top-$k$ eviction during decode. We also derive a coreset-theoretic justification (Sec.~\ref{sec:framework_theory}) showing why the same ranking serves both phases.
\item \textbf{Three-axis Pareto study (Sec.~\ref{sec:exp_prefill} and Sec.~\ref{sec:exp_decode}):} across seven model variants and three full-pipeline-probe benchmarks, density-based prefill compression matches the strongest accuracy frontier while scaling linearly in the visual-token count and selecting tokens an order of magnitude faster at high resolution.
\item \textbf{Cross-architecture and cross-scale robustness (Appendix~\ref*{app:robustness}):} a structural risk taxonomy explaining why pure-selection methods catastrophically fail on Qwen3-VL-4B's multi-axis RoPE while merge-based density does not, together with the empirical finding that compression cost \emph{does not grow} with model size across the InternVL-3.5 from 2B to 38B.
\end{enumerate}
\vspace{-2mm}
\section{Related Work}\label{sec:related}
\vspace{-2mm}
\paragraph{Prefill-stage compression.} FastV~\cite{chen2024fastv} drops visual tokens by attention score after a chosen layer. ToMe~\cite{bolya2023tome} performs bipartite token merging for ViTs and was adapted to VLMs. DivPrune~\cite{divprune2025} uses farthest-point sampling on visual features for diversity-driven selection. DART~\cite{dart2025} combines L1-norm pivots with cosine-distance expansion. VisionZip~\cite{visionzip2025} selects dominant tokens by CLS-attention and merges the rest by key-state similarity. TokenPacker~\cite{tokenpacker2025} introduces a coarse-to-fine visual projector that compresses tokens by 75 to 89 percent before the LLM backbone. AvgPool, FitPrune~\cite{fitprune}, and PruMerge~\cite{prumerge} round out a crowded landscape, and concurrent preprints continue to appear, including SAINT~\cite{saint}. All evaluate primarily on the four-benchmark set we critique below.

\paragraph{Decode-stage KV eviction.} For text-only LLMs, H2O~\cite{h2o2023} and StreamingLLM~\cite{streamingllm} maintain ``heavy hitter'' KV entries. SnapKV~\cite{snapkv} aggregates attention. FastGen~\cite{fastgen2024} profiles attention-head patterns to build a query-adaptive eviction policy. Quest~\cite{quest2024} selects KV pages by query similarity for efficient long-context inference. KV quantization, exemplified by the 2-bit asymmetric KIVI~\cite{kivi2024}, provides a complementary compression axis orthogonal to eviction. On the visual side, LOOK-M~\cite{lookm2024} merges multimodal KV entries guided by text-prior attention. ElasticCache~\cite{elasticcache2024} uses importance-driven KV merging for vision instruction-following models. R-KV~\cite{rkv2025} combines attention aggregation with a redundancy penalty for reasoning-LLM KV compression. Most works report on the same prefill-probe benchmarks that, as we show, cannot measure their decode-phase effect.

\paragraph{Two-phase view.} To our knowledge, no prior work explicitly separates prefill (compute) and decode (memory) as orthogonal axes with distinct algorithms and evaluation requirements. Our framework makes this separation a first-class design principle.

\paragraph{Token pruning as coreset selection.} Keeping $k$ of $N$ visual tokens is a \emph{coreset} query on $X \in \mathbb{R}^{N\times d}$, with two dual classical strategies: $k$-center diversity (DivPrune~\cite{divprune2025}, Sener and Savarese~\cite{senerSavarese2018}) and density- or sensitivity-weighted sampling (Feldman and Langberg~\cite{feldmanLangberg2011}, Bachem, Lucic, and Krause~\cite{bachemLucicKrause2017}, submodular maximization~\cite{mirzasoleiman2015}). Our density signal $\rho$ is a training-free, deterministic-threshold proxy of the latter family (Appendix~\ref*{app:coreset}). We are not aware of prior work framing VLM token pruning as a \emph{two-phase} coreset whose subset must simultaneously approximate prefill attention and decode KV recall, and this paper is the first to exploit the resulting signal reuse.

\section{Two-Phase Compression, Two-Phase Diagnostic}\label{sec:methodology}

\subsection{The argmax mechanism}

Consider a VLM prefill that yields hidden states $H \in \mathbb{R}^{T \times d}$ over $T$ tokens (visual + text). The first decoded token is
\begin{equation}
y_1 = \arg\max_v \big[ W_o\,h_T \big]_v,
\label{eq:argmax}
\end{equation}
where $h_T$ is the last hidden state and $W_o$ is the language-model head. Decode-phase compression, for example KV eviction $\mathcal{E}$ acting on the cached $\{(K_\ell, V_\ell)\}$, takes effect only when computing $y_2, y_3, \ldots$ from the now-reduced cache. For benchmarks whose ground-truth answer is a single token, such as a multiple-choice letter or a yes/no token, the score depends only on $y_1$. Thus decode-phase $\mathcal{E}$ has no effect on these scores by construction. Density-based compression operates on \emph{both} prefill and decode, with measurable gains in each phase (Sec.~\ref{sec:exp_prefill} and Sec.~\ref{sec:exp_decode}). To verify the two phases independently, we partition benchmarks into two diagnostic roles based on their scoring rule. A benchmark is a \emph{prefill probe} if its score depends only on $y_1$, as in a multiple-choice letter, a yes/no token, or any single-letter argmax over a fixed set such as MMBench, POPE, and ScienceQA. Such a probe isolates prefill quality and is structurally blind to any decode-phase change. A benchmark is a \emph{full-pipeline probe} if its score depends on the decoded sequence $y_1, y_2, \ldots$, including free-form answers scored by exact-match, ANLS edit distance, or an LLM judge such as TextVQA, DocVQA, and MM-Vet. Such a probe reveals both phases and their interaction effects. The split is a property of the scoring rule, not the literal answer length during generation.
\vspace{-2mm}
\subsection{Empirical proof}
\vspace{-2mm}
We instantiate $\mathcal{E}$ ranging from \textsc{D75} (keep $25\%$ of visual KV) to \textsc{Dall} (drop \emph{all} visual KV). Table~\ref{tab:mismatch} reports prediction mismatch \% relative to the no-compression baseline.

\begin{table}[t]
\caption{Prediction-mismatch \% vs.\ no-compression baseline on InternVL-3.5-8B. Decode KV eviction flips up to $46\%$ of TextVQA predictions but $\le 0.54\%$ on prefill-probe benches, since $y_1$ depends only on prefill logits (Eq.~\ref{eq:argmax}). Prefill rows fix $L{=}15$ (varying $L$ shifts levels but not the scoring-rule gap). Appendix~\ref*{app:mismatch} extends to all 3 archs.}
\label{tab:mismatch}
\centering
\small
\setlength{\tabcolsep}{3mm}
\resizebox{\linewidth}{!}{
\begin{tabular}{llcccc}
\toprule
Phase & Signal · Keep & MMBench & POPE & SQA & TextVQA \\
\midrule
Prefill & Density · $75$/$50$/$25\%$  & $0$    & $0$--$0.03$ & --   & $3.1$/$7.0$/$14.9$ \\
Prefill & Random · $25\%$             & $0$    & $0$         & --   & $32.8$ \\
Prefill & Attention · $25\%$          & $0$    & $0.03$      & --   & $25.1$ \\
\midrule
Decode  & Density · $25$/$12.5$/$6.25\%$ & $0.54$ & $0.32$--$0.35$ & $0.20$ & $15.9$/$23.0$/$31.1$ \\
\textbf{Decode}  & \textbf{Density · $0\%$ (drop all)}      & \textbf{0.54} & \textbf{0.38} & \textbf{0.20} & \textbf{45.5} \\
\bottomrule
\end{tabular}}
\vspace{-2mm}
\end{table}

\paragraph{Implication.} The pattern is universal across architectures: TextVQA's $45.52\%$ mismatch under \textsc{Dall} (Tab.~\ref{tab:mismatch}) refutes the ``tied logits'' objection, and Appendix~\ref*{app:mismatch} extends the cross-model grid to confirm full-pipeline probes register both phases while prefill probes register neither in practice. Any claim of decode-phase token pruning that preserves accuracy validated only on MMBench, POPE, or ScienceQA must be re-examined; future VLM token-pruning evaluation should include at least one full-pipeline-probe benchmark such as MM-Vet, DocVQA, or LLaVA-Bench-Wilder.

\section{Density-Guided Two-Phase Compression}\label{sec:framework}
\begin{figure}[t]
    \centering
    \includegraphics[width=\linewidth]{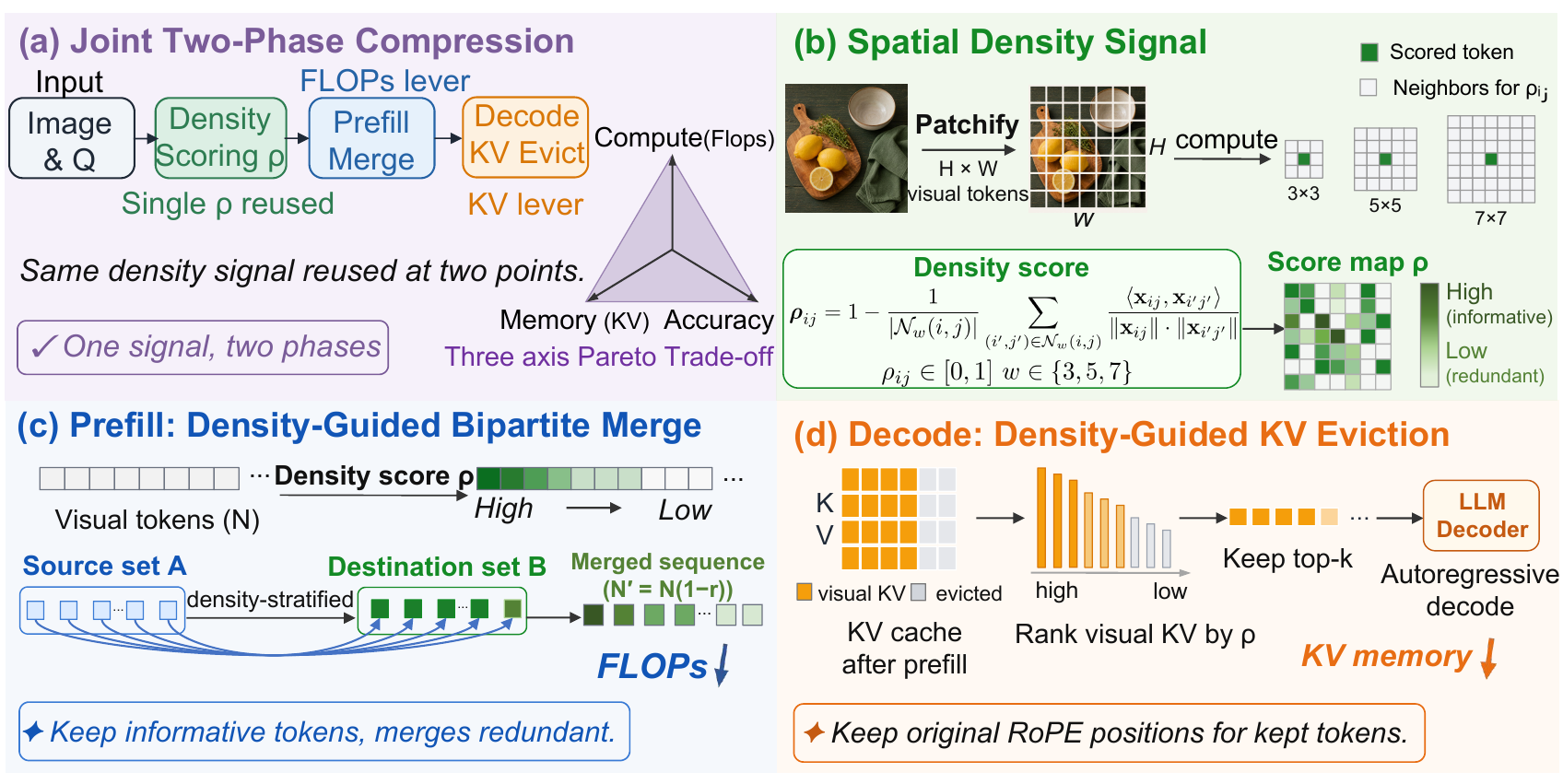}
    \caption{\textbf{\ours{} pipeline.} A single spatial density signal $\rho$, defined as each visual token's average cosine dissimilarity to its $w{\times}w$ neighborhood (b), drives both phases. \textbf{Prefill} bipartite-merges low-$\rho$ source tokens into high-$\rho$ destination anchors, cutting FLOPs (c). \textbf{Decode} ranks visual KV by $\rho$ and keeps the top-$k$ with original RoPE positions, cutting KV memory (d). One signal, two levers, and a three-axis Pareto trade-off over compute, KV, and accuracy (a).}
    \label{fig:method}
\end{figure}
\subsection{Spatial density signal}

For visual tokens arranged on a grid of size $H \times W$, define density at position $(i,j)$ via local cosine dissimilarity:
\begin{equation}
\rho_{ij} = 1 - \frac{1}{|\mathcal{N}_w(i,j)|} \sum_{(i',j') \in \mathcal{N}_w(i,j)} \frac{\langle x_{ij}, x_{i'j'} \rangle}{\|x_{ij}\| \cdot \|x_{i'j'}\|},
\label{eq:density}
\end{equation}
where $\mathcal{N}_w(i,j)$ is the spatial $w \times w$ neighborhood (excluding self). We compute $\rho$ at multiple scales $w \in \{3, 5, 7\}$ and take the max, capturing both fine and coarse structure. High $\rho_{ij}$ indicates that the token differs from its neighbors and is therefore informative, whereas low $\rho_{ij}$ indicates redundancy.

\paragraph{Why density?} Density is an \emph{intrinsic, query-invariant} feature-tensor statistic. It is monotone in a finite-sample kernel-$H_2$ estimator~\cite{renyi1961}, contrasting the luminance-biased $\ell_2$-norm and the sink-biased, query-dependent attention scores. Section~\ref{sec:signal_ablation} confirms density wins or ties at prefill across all three probe benchmarks and wins on the two deterministic full-pipeline probes at decode (information-geometric framing and the failed QC-density extension are in Appendix~\ref*{app:qc_drift}).

\paragraph{Coreset reading.}\label{sec:framework_theory} A coreset that approximates the visual-token feature matrix approximates every downstream computation that accesses it through attention aggregates, so prefill merge and decode eviction inherit the \emph{same} approximation guarantee from a single density top-$k$ subset. Theorem~1 in Appendix~\ref*{app:coreset} bounds the approximation error by $\varepsilon \leq 2eQr_k/\sqrt{d}$ where $r_k$ is the $k$-center cover radius of the chosen subset. Density top-$k$ minimises $r_k$ in expectation while farthest-point sampling minimises it by construction, which explains why the two greedy heuristics agree at mild compression and diverge at aggressive compression on skewed distributions (Sec.~\ref{sec:exp_prefill}). A Feldman--Langberg-style $1/\sqrt{k}$ rate (Theorem~2) and the full proofs are in Appendix~\ref*{app:coreset}.

\subsection{Prefill: density-tilted farthest-point selection}\label{sec:method_prefill}

At a chosen insertion layer $L$ during prefill, we keep $k = \lceil N \cdot (1{-}r) \rceil$ of the $N$ visual tokens by a density-tilted greedy selection. Compute density $\rho$ on the layer-$L$ hidden states via Eq.~\ref{eq:density}, normalize to $\tilde\rho \in [0,1]$, and pre-compute the pairwise cosine distance matrix $D_{ij}=1-\langle \hat x_i,\hat x_j\rangle$ where $\hat x = x/\|x\|$. Seed with $s_0 = \arg\max_i \rho_i$, the most informative token, and maintain $\mathrm{mind}_j = D_{s_0 j}$ as the minimum distance from $j$ to the selected set. For each subsequent step, pick
\begin{equation}
s_t \;=\; \arg\max_{j \notin S} \big(\,\mathrm{mind}_j \;+\; \beta\,\tilde\rho_j\,\big),
\quad\text{then}\quad \mathrm{mind}_j \leftarrow \min(\mathrm{mind}_j,\,D_{s_t j}),
\label{eq:fps_density_tilt}
\end{equation}
with $\beta = 0.3$ throughout. Selected indices are sorted to preserve the visual-token-grid ordering, and KV-cache rows for the unselected tokens are dropped before the next layer. Eq.~\ref{eq:fps_density_tilt} unifies two greedy heuristics for the same coreset, as discussed in Sec.~\ref{sec:framework_theory}. Pure farthest-point sampling at $\beta{=}0$ recovers DivPrune~\cite{divprune2025}, and density top-$k$ is the limit $\beta{\to}\infty$. The small additive $\beta=0.3$ tilt lets coverage dominate at high prune ratios while density breaks ties when $\mathrm{mind}_j$ saturates, which is the regime where pure FPS picks isolated outliers with low local relevance.

\paragraph{Implementation.} The $k$-step inner loop is the latency-critical path of density-tilted FPS. A naive cuBLAS-plus-graph realisation issues five kernel launches per step, and at our operating point ($N{=}3072$, $D{=}4096$, $K{=}614$) the per-launch overhead dominates the actual compute. We fuse the body into two Triton kernels, an argmax reduction and an inner-iter $\mathrm{mind}$ update with a $32\!\times\!64$ row-column tile, and back the recurrence with an fp16 normalized-token buffer. fp16's 10-bit mantissa resolves $10^{-3}$ differences in cosine similarity, comfortably finer than the random-vector cosine standard deviation $1/\sqrt{D}{=}0.016$ at $D{=}4096$. The full recurrence is captured into a CUDA graph keyed on $(N, D, k, \beta)$, replayed once per layer rather than re-launched per step, and cached at module scope across the sweep. Without this cache, the per-call capture overhead of roughly $80$\,ms inverts the legacy-versus-graph latency ranking at our $K$. The combination delivers a $3.24\times$ end-to-end prefill speedup at the B3 cell on InternVL-3.5-8B, with cached-Triton fp16 \ours{} at $150.8$\,ms per sample versus the $487.7$\,ms uncompressed baseline.

\subsection{Decode: density-guided KV eviction}\label{sec:method_decode}

After prefill completes, with or without prefill-stage compression, evict visual KV entries by density rank in a single shot. We keep $k_{\text{dec}} = \lceil N_{\text{post}} \cdot r_{\text{decode}} \rceil$ entries with the highest $\rho$, where $N_{\text{post}}$ is the post-prefill visual-token count. Surviving KV rows retain their original RoPE-rotated keys without rotation re-application, and the first decoded token is positioned at $\text{original\_prefill\_length}{+}1$ to maintain query and key alignment. Density is reused from prefill when both phases are active, so the decode stage adds no extra similarity computation. Eviction is one-shot at the prefill-decode boundary, in contrast to streaming approaches that maintain a sliding window during generation. The choice is justified by the coreset reading in Sec.~\ref{sec:framework_theory}: a density-coreset of the post-prefill visual tokens approximates every subsequent attention aggregate that queries those tokens.

\subsection{Joint and the three-axis Pareto}

Joint compression composes prefill merge with decode eviction. The resulting design space is naturally three-dimensional: (i) prefill FLOPs (reduced by merge), (ii) KV memory (reduced by eviction), (iii) accuracy. Single-phase methods occupy the axes, and joint configurations fill the interior. We characterize this Pareto surface in Sections~\ref{sec:exp_prefill} and~\ref{sec:exp_decode}, and Algorithm~\ref*{alg:unified_density} in Appendix~\ref*{app:algorithm} gives the full pseudocode.


\begin{table}[t]
\caption{Per-benchmark prefill compression on \textbf{InternVL-3.5-8B}; same $(L,r)$ within each block. Bold marks per-column best; daggers (\dag) flag structurally invalid output (acc\,$\approx\!0$, App.~\ref*{app:crossarch_fragility}). Block 4 in App.~\ref*{app:block4_high_tpl}. \textbf{Speedup}: measured prefill speedup vs.\ uncompressed baseline at batch$=$8, HD mode (12 tiles, $N_{\text{vis}}{=}3072$) on RTX A6000.}
\label{tab:prefill_internvl}
\centering
\small
\setlength{\tabcolsep}{3pt}
\begin{tabular}{llccccccc}
\toprule
Method & Venue & TPL\,\% & MMBench & POPE & SQA & TextVQA & Avg-4 & Speedup \\
\midrule
\multicolumn{2}{l}{\textit{Baseline} (no compression)}                 & 100.0 & 0.8707 & 0.8903 & 0.9539 & 0.6493 & 0.8411 & $1.00\times$ \\
\midrule
\multicolumn{9}{l}{\textit{Block 1: low-TPL — $L\!=\!15$, prune $50\%$, TPL\,$=65.2\%$}} \\
Random            & --                              & 65.2 & 0.8862          & \textbf{0.9155} & 0.9782          & 0.6202          & 0.8500          & $\mathbf{1.28\times}$ \\
ToMe              & ICLR'23~\cite{bolya2023tome}    & 65.2 & 0.8584          & 0.8913          & 0.9346          & 0.5325          & 0.8042          & $1.16\times$ \\
FastV             & ECCV'24~\cite{chen2024fastv}    & 65.2 & 0.8584          & 0.8868          & 0.9326          & 0.5150          & 0.7982          & $\mathbf{1.28\times}$ \\
PruMerge          & ICCV'25~\cite{prumerge}         & 65.2 & 0.8584          & 0.8943          & 0.9410          & 0.5937          & 0.8219          & $1.21\times$ \\
DivPrune          & CVPR'25~\cite{divprune2025}     & 65.2 & 0.8630          & 0.8923          & 0.9430          & 0.6017          & 0.8250          & $1.17\times$ \\
DART              & EMNLP'25~\cite{dart2025}        & 65.2 & 0.8467          & 0.8877          & 0.9316          & 0.6026          & 0.8171          & $\mathbf{1.28\times}$ \\
CAC (ours)        & --                              & 65.2 & \textbf{0.8909} & 0.9142          & \textbf{0.9792} & \textbf{0.7195} & \textbf{0.8759} & $1.16\times$ \\
\midrule
\multicolumn{9}{l}{\textit{Block 2: extreme stress — $L\!=\!3$, prune $85\%$, TPL\,$=24.1\%$}} \\
Random            & --       & 24.1 & 0.7980          & 0.8398          & 0.8622          & 0.3285          & 0.7071          & $\mathbf{3.12\times}$ \\
ToMe              & ICLR'23  & 24.1 & \textbf{0.8320} & \textbf{0.8798} & 0.8860          & 0.4524          & 0.7626          & $2.07\times$ \\
FastV             & ECCV'24  & 24.1 & 0.7423          & 0.7665          & 0.7833          & 0.1939          & 0.6215          & $3.11\times$ \\
PruMerge          & ICCV'25  & 24.1 & 0.8313          & 0.8578          & 0.8825          & 0.3866          & 0.7396          & $2.47\times$ \\
DivPrune\dag      & CVPR'25  & 24.1 & 0.0000          & 0.0000          & 0.0000          & 0.0041          & 0.0010          & $2.98\times$ \\
DART\dag          & EMNLP'25 & 24.1 & 0.0000          & 0.0000          & 0.0000          & 0.0041          & 0.0010          & $3.09\times$ \\
CAC (ours)        & --       & 24.1 & 0.8282          & 0.8670          & \textbf{0.8929} & \textbf{0.5696} & \textbf{0.7894} & $2.67\times$ \\
\midrule
\multicolumn{9}{l}{\textit{Block 3: visual-only pure prune — $L\!=\!0$, prune $80\%$, TPL\,$=20.0\%$}} \\
Random            & --       & 20.0 & 0.8297          & 0.8565          & 0.8751          & 0.3665          & 0.7320          & $3.57\times$ \\
ToMe              & ICLR'23  & 20.0 & 0.8351          & 0.8893          & 0.8875          & 0.4599          & 0.7679          & $2.31\times$ \\
FastV             & ECCV'24  & 20.0 & 0.8320          & 0.8057          & \textbf{0.9043} & \textbf{0.6149} & 0.7892          & $\mathbf{3.58\times}$ \\
PruMerge          & ICCV'25  & 20.0 & 0.8297          & 0.8715          & 0.8875          & 0.3399          & 0.7321          & $2.79\times$ \\
DivPrune\dag      & CVPR'25  & 20.0 & 0.0000          & 0.0000          & 0.0000          & 0.0041          & 0.0010          & $3.11\times$ \\
DART\dag          & EMNLP'25 & 20.0 & 0.0000          & 0.0000          & 0.0000          & 0.0041          & 0.0010          & $3.30\times$ \\
CAC (ours)        & --       & 20.0 & 0.8467          & 0.8835          & 0.8914          & 0.5793          & 0.8003          & $3.24\times$ \\
\midrule
\multicolumn{9}{l}{\textit{Block 3b: + LoRA recovery, fair control — $L\!=\!0$, prune $70\%$, TPL\,$=30.0\%$ (matched)}} \\
Random + LoRA     & --       & 30.0 & 0.8615          & 0.8983          & 0.9142          & 0.4512          & 0.7813          & $1.32\times$ \\
DivPrune + LoRA   & CVPR'25  & 30.0 & \textbf{0.8607} & \textbf{0.9060} & \textbf{0.9355} & 0.5981          & 0.8251          & $0.92\times$ \\
DART + LoRA       & EMNLP'25 & 30.0 & 0.8475          & 0.8837          & 0.9346          & 0.6149          & 0.8202          & $1.83\times$ \\
CAC + LoRA (ours) & --       & 30.0 & 0.8514          & 0.9058          & 0.9321          & \textbf{0.6385} & \textbf{0.8319} & $\mathbf{2.14\times}$ \\
\bottomrule
\end{tabular}
\end{table}

\section{Prefill compression Pareto}\label{sec:exp_prefill}

\subsection{Setup}\label{sec:exp_setup}

We evaluate seven model variants from three VLM families: InternVL-3.5 at 2B, 4B, 8B, and 38B~\cite{internvl35}, Qwen3-VL at 2B and 4B~\cite{qwen3vl}, and LLaVA-1.5-7B~\cite{llava15}. Prefill-probe benchmarks (MMBench~\cite{mmbench}, POPE~\cite{pope}, ScienceQA~\cite{scienceqa}) score the first decoded token, isolating prefill quality. Full-pipeline-probe benchmarks reveal both phases. The full-pipeline-probe set comprises TextVQA~\cite{textvqa} ($n{=}5000$, exact-match), MM-Vet~\cite{mmvet} ($n{=}218$, free-form, official MM-Vet evaluator), and DocVQA ($n{=}5349$, deterministic ANLS scoring). On the prefill side we compare density-based merging (\ours{} and a variant \ours{}-2a, ours) against DivPrune~\cite{divprune2025}, DART~\cite{dart2025}, FastV~\cite{chen2024fastv}, ToMe~\cite{bolya2023tome}, and PruMerge~\cite{prumerge}. On the decode side, we compare density (ours) against random and attention-value-norm baselines. We sweep prefill prune $\in\{25, 50, 75\}\%$ at insertion layers $\{0,1,2,3,5,10,15,20\}$ and decode keep $\in\{75, 50, 25, 12.5, 6.25, 0\}\%$.

\paragraph{Configuration shorthand.} We refer to specific compression configurations by short codes throughout. \texttt{\ours\_L$L$\_$r$} denotes prefill-only \ours{} compression inserted at LLM layer $L$ with visual-token prune ratio $r\%$, so \texttt{\ours\_L3\_50} is prefill-only compression at layer 3 pruning $50\%$ of visual tokens. \texttt{decode\_only\_D$d$} denotes decode-only KV eviction that drops $d\%$ of the visual KV span after prefill, so \texttt{D75} keeps $25\%$ of visual KV, \texttt{D875} keeps $12.5\%$, and \texttt{Dall} drops every visual KV. \texttt{joint\_P$p$\_D$d$} denotes joint compression that combines prefill-merge prune $p\%$ at layer 3 with decode-eviction keep $(100{-}d)\%$, so \texttt{joint\_P50\_D75} is layer-3 prefill prune $50\%$ paired with decode keep $25\%$.

\subsection{Prefill compression Pareto}


Table~\ref{tab:prefill_internvl} reports per-benchmark accuracies on InternVL-3.5-8B under a fair-comparison protocol where each block fixes a single $(L, r)$ pair shared by all methods, isolating the selection-rule difference. Three blocks in the main paper span the TPL spectrum. Block B1 uses $L\!=\!15$ with prune $50\%$ and TPL$\approx65\%$. Block B2 uses $L\!=\!3$ with prune $85\%$ and TPL$\approx24\%$, an extreme stress setting. Block B3 uses $L\!=\!0$ with prune $80\%$ and TPL$\approx20\%$, a visual-only pure-prune setting. The saturated high-TPL block B4 at $L\!=\!20$, prune $50\%$, TPL$\approx86\%$ is reported in Appendix~\ref*{app:block4_high_tpl}. The matched cross-architecture sweep on Qwen3-VL-4B uses the same four-block schedule with M-RoPE 3-axis position encoding and is reported in Appendix~\ref*{app:prefill_qwen} as Tab.~\ref*{tab:prefill_qwen}.

\paragraph{Headline.} \ours{} leads Avg-4 in every block on InternVL-8B. The lead is $+2.6$pp at the moderate B1 ($L\!=\!15$, prune $50\%$) and grows to $+1.1$pp over the next \emph{healthy} method at the visual-only B3 ($L\!=\!0$, prune $80\%$), where pure-selection methods DivPrune and DART collapse to acc\,$\approx\!0$. Under the fair LoRA control at matched TPL$=30\%$ (B3b), \ours{}+LoRA reaches Avg-4 $=0.8319$ vs DivPrune+LoRA at $0.8251$, with a $+4.0$pp gap on TextVQA. Block-by-block reading and the matched Qwen3-VL-4B sweep are in Appendix~\ref*{app:prefill_reading}. Selection-side complexity is in Appendix~\ref*{app:complexity}.

\subsection{Signal ablation}\label{sec:signal_ablation}

Is the advantage of our method really density, or could a simpler score match it? Holding every pipeline component fixed ($L{=}15$, prune $50\%$, InternVL-3.5-8B, prune-only top-$k$) and swapping only the ranking score across four benchmarks (TextVQA with $n{=}5000$, MMBench, V$^{\ast}$Bench, MM-Vet), density wins or ties on all four: it outperforms L2 norm on TextVQA by 2.2 percentage points ($0.616$ vs.\ $0.594$), attention V-norm on TextVQA by 3.2 percentage points, and V-norm on V$^{\ast}$Bench by 4.7 percentage points. MMBench scores cluster near $0.86$ because the prefill-probe scoring rule is insensitive to which visual tokens are kept (Sec.~\ref{sec:methodology}). Multi-scale density is nearly indistinguishable from single-scale, differing by 0.03 percentage points on TextVQA, and the product $\rho\!\cdot\!\|v\|$ is neutral-to-hurt: the two signals are not independent, so combining them injects noise. Full four-benchmark table in Appendix~\ref*{app:extended_baselines}.

\section{Decode compression on full-pipeline-probe benchmarks}\label{sec:exp_decode}

\begin{figure}[t]
\centering
\includegraphics[width=\linewidth]{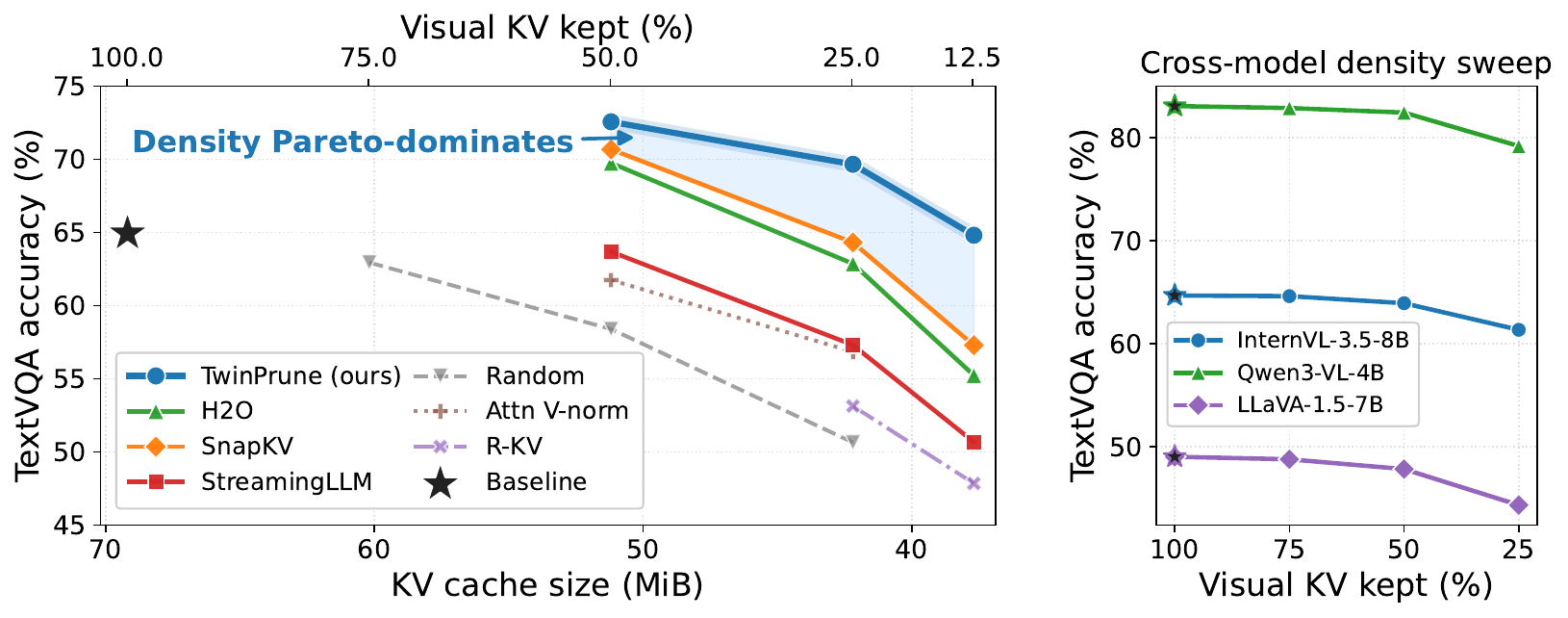}
\caption{\textbf{Decode KV Eviction on TextVQA.}
On InternVL-3.5-8B, density-guided eviction Pareto-dominates random and attention-value-norm in the memory--accuracy trade-off.
Right inset: Qwen3-VL-4B, LLaVA-1.5-7B, and InternVL-3.5-38B show the same monotonic density trend.}
\label{fig:decode}
\end{figure}

\subsection{Decode KV eviction Pareto (Figure~\ref{fig:decode})}\label{sec:exp_decode_pareto}

Density-guided eviction outperforms random selection by 5 to 11 percentage points at aggressive ($75\%$) eviction and outperforms the value-norm proxy by 4 to 8 percentage points, consistently across all six tested models. The signal is robust under cross-architecture variation and, as Section~\ref{sec:signal_ablation} shows, outperforms L2-norm by 2.2 percentage points on TextVQA at matched configuration.

\paragraph{Comparison with published decode-eviction baselines.} Table~\ref{tab:decode_methods} extends the comparison to H2O~\cite{h2o2023}, SnapKV~\cite{snapkv}, and StreamingLLM~\cite{streamingllm} on the visual-KV span at three keep ratios. Density top-$k$ wins all six TextVQA and MM-Vet cells outright, and the advantage \emph{widens} at extreme compression: at keep $=12.5\%$ density beats H2O by $9.6$, $5.4$, and $3.5$pp on TextVQA, MM-Vet, and DocVQA respectively. SnapKV narrowly leads density on DocVQA at the looser keeps but density reclaims the lead at $12.5\%$. V-norm takes MM-Vet at decode by $4.8$pp at keep $25\%$ (Tab.~\ref*{tab:decode_signal_ablation}); density still wins TextVQA by $4.5$pp and DocVQA by $2.4$pp at the same keep, and V-norm requires decode-time attention maps that break the two-phase signal reuse (Appendix~\ref*{app:extended_baselines}).

\begin{table}[t]
\caption{Decode-phase KV eviction comparison on InternVL-3.5-8B at three keep ratios. Density top-$k$ wins all six TextVQA and MM-Vet cells. On DocVQA, SnapKV narrowly leads at keep $\ge 25\%$ but density reclaims the lead at $12.5\%$ keep. Bold marks per-column best. \textbf{TextVQA} uses $n{=}5000$ exact-match. \textbf{MM-Vet} uses $n{=}218$. \textbf{DocVQA} uses $n{=}5349$ with ANLS scoring.}
\label{tab:decode_methods}
\centering
\small
\setlength{\tabcolsep}{2mm}
\resizebox{\linewidth}{!}{
\begin{tabular}{l ccc ccc ccc}
\toprule
& \multicolumn{3}{c}{TextVQA} & \multicolumn{3}{c}{MM-Vet} & \multicolumn{3}{c}{DocVQA} \\
\cmidrule(lr){2-4}\cmidrule(lr){5-7}\cmidrule(lr){8-10}
Method (keep $k$) & $50\%$ & $25\%$ & $12.5\%$ & $50\%$ & $25\%$ & $12.5\%$ & $50\%$ & $25\%$ & $12.5\%$ \\
\midrule
Baseline (no eviction) & \multicolumn{3}{c}{$0.733$} & \multicolumn{3}{c}{$0.756$} & \multicolumn{3}{c}{$0.642$} \\
\midrule
StreamingLLM~\cite{streamingllm}  & 0.637          & 0.573          & 0.507          & 0.667          & 0.561          & 0.514          & 0.468          & 0.384          & 0.324          \\
H2O~\cite{h2o2023}                & 0.697          & 0.629          & 0.552          & 0.662          & 0.571          & 0.497          & 0.555          & 0.438          & 0.347          \\
SnapKV~\cite{snapkv}              & 0.707          & 0.643          & 0.573          & 0.654          & 0.585          & 0.513          & \textbf{0.577} & \textbf{0.469} & 0.377          \\
Density (ours)                    & \textbf{0.726} & \textbf{0.697} & \textbf{0.648} & \textbf{0.708} & \textbf{0.634} & \textbf{0.551} & 0.575          & 0.466          & \textbf{0.382} \\
\bottomrule
\end{tabular}}
\end{table}
\vspace{-2mm}
\subsection{Joint compression analysis}\label{sec:exp_joint}
\vspace{-2mm}
Per-model MM-Vet scores under \texttt{\ours\_L3\_50}, \texttt{decode\_only\_D75}, and \texttt{joint\_P50\_D50} are in Tab.~\ref*{tab:mmvet} (Appendix~\ref*{app:mmvet}); both single phases cost $10$ to $18$pp, with no free lunch. The 3$\times$3 dense grid (Tab.~\ref*{tab:joint_interior_textvqa}, Appendix~\ref*{app:joint_mmvet}) confirms a smooth bilinear surface in the interior: accuracy falls monotonically along both axes, and joint configurations are legitimate budget trades but never Pareto-dominate single-phase methods at matched KV budgets. A complementary 3-prefill$\times$3-decode-signal sweep (Appendix~\ref*{app:joint_sweep}) shows density's lead over H2O is concentrated in the decode-only regime; once aggressive prefill at $L{=}3,r{=}85\%$ has removed the redundant visual signal, the three decode signals collapse within $1$pp on TextVQA.

\vspace{-2mm}
\section{Discussion and Conclusion}\label{sec:discussion}
\vspace{-2mm}

\paragraph{Summary and deployment.} VLM token pruning has been evaluated on a benchmark mix that limits visibility into decode-phase effects. We provide the proof, propose a unified density-based framework that scores token importance for both phases, and supply a three-axis Pareto across seven model variants and three full-pipeline-probe benchmarks. We further identify a structural risk class explaining when pure-selection methods fail on multi-axis RoPE backbones, and we show compression cost \emph{does not grow} with model size in the 8B-to-38B regime. In deployment, latency-bound use cases use prefill compression ($3.24\times$ speedup at the B3 cell), memory-bound use cases use decode eviction, and joint configurations apply when both axes are constrained.\footnote{Compute: NVIDIA A6000 (48\,GB) cluster, with the 38B model sharded across two A6000s. The reported experiments consumed approximately 2{,}000 to 3{,}000 GPU-hours, and exploratory runs (Appendix~\ref*{app:qc_drift}) add a comparable unreported budget. Code, predictions, and scripts will be released upon publication.}

\vspace{-2mm}
\paragraph{Limitations.} We identify two principled limitations. (i)~Single-stage prefill compression, with multi-stage scheduling unexplored. (ii)~Spatial-grid density assumes regular tokenization, so non-grid tokenizers such as Q-Former require alternative formulations.

\vspace{-2mm}
\paragraph{Broader impact and future work.} Efficient VLM serving lowers energy use and enables edge deployment. Compression can subtly alter behavior on long-tail or safety-critical inputs without changing prefill-probe scores, so high-stakes deployments must validate on full-pipeline-probe benchmarks. Future work: learning the optimal density-aggregation kernel, and task-conditioned signals beyond pre-attention geometry (Appendix~\ref*{app:qc_drift}).


\clearpage
\begin{thebibliography}{10}

\bibitem{divprune2025}
Saeed~Ranjbar Alvar, Gursimran Singh, Mohammad Akbari, and Yong Zhang.
\newblock {DivPrune}: Diversity-based visual token pruning for large multimodal
  models.
\newblock In {\em CVPR}, 2025.
\newblock arXiv:2503.02175.

\bibitem{bachemLucicKrause2017}
Olivier Bachem, Mario Lucic, and Andreas Krause.
\newblock Practical coreset constructions for machine learning.
\newblock {\em arXiv preprint arXiv:1703.06476}, 2017.

\bibitem{qwen3vl}
Shuai Bai, Yuxuan Cai, Ruizhe Chen, Keqin Chen, Xionghui Chen, Zesen Cheng,
  Lianghao Deng, Wei Ding, Chang Gao, Chunjiang Ge, et~al.
\newblock Qwen3-vl technical report.
\newblock {\em arXiv preprint arXiv:2511.21631}, 2025.

\bibitem{bolya2023tome}
Daniel Bolya, Cheng-Yang Fu, Xiaoliang Dai, Peizhao Zhang, Christoph
  Feichtenhofer, and Judy Hoffman.
\newblock Token merging: Your {ViT} but faster.
\newblock In {\em ICLR}, 2023.

\bibitem{rkv2025}
Zefan Cai, Wen Xiao, Hanshi Sun, Cheng Luo, Yikai Zhang, Ke~Wan, Yucheng Li,
  Yeyang Zhou, Li-Wen Chang, Jiuxiang Gu, et~al.
\newblock R-kv: Redundancy-aware kv cache compression for reasoning models.
\newblock In {\em NeurIPS}, 2025.

\bibitem{chen2024fastv}
Liang Chen, Haozhe Zhao, Tianyu Liu, Shuai Bai, Junyang Lin, Chang Zhou, and
  Baobao Chang.
\newblock An image is worth 1/2 tokens after layer 2: Plug-and-play inference
  acceleration for large vision-language models.
\newblock In {\em ECCV}, 2024.

\bibitem{feldmanLangberg2011}
Dan Feldman and Michael Langberg.
\newblock A unified framework for approximating and clustering data.
\newblock {\em Proceedings of the 43rd ACM Symposium on Theory of Computing
  (STOC)}, pages 569--578, 2011.
\newblock arXiv:1106.1379.

\bibitem{fastgen2024}
Suyu Ge, Yunan Zhang, Liyuan Liu, Minjia Zhang, Jiawei Han, and Jianfeng Gao.
\newblock Model tells you what to discard: Adaptive {KV} cache compression for
  {LLMs}.
\newblock In {\em ICLR}, 2024.

\bibitem{saint}
Ahmadreza Jeddi, Negin Baghbanzadeh, Elham Dolatabadi, and Babak Taati.
\newblock Similarity-aware token pruning: Your {VLM} but faster.
\newblock {\em arXiv preprint arXiv:2503.11549}, 2025.

\bibitem{langbergSchulman2010}
Michael Langberg and Leonard~J. Schulman.
\newblock Universal $\varepsilon$-approximators for integrals.
\newblock In {\em Proceedings of the 21st Annual ACM-SIAM Symposium on Discrete
  Algorithms (SODA)}, pages 598--607, 2010.

\bibitem{tokenpacker2025}
Wentong Li, Yuqian Yuan, Jian Liu, Dongqi Tang, Song Wang, Jie Qin, Jianke Zhu,
  and Lei Zhang.
\newblock {TokenPacker}: Efficient visual projector for multimodal {LLM}.
\newblock {\em International Journal of Computer Vision}, 133:6794--6812, 2025.

\bibitem{pope}
Yifan Li, Yifan Du, Kun Zhou, Jinpeng Wang, Xin Zhao, and Ji-Rong Wen.
\newblock Evaluating object hallucination in large vision-language models.
\newblock In {\em EMNLP}, 2023.

\bibitem{snapkv}
Yuhong Li, Yingbing Huang, Bowen Yang, Bharat Venkitesh, Acyr Locatelli,
  Hanchen Ye, Tianle Cai, Patrick Lewis, and Deming Chen.
\newblock Snapkv: Llm knows what you are looking for before generation.
\newblock {\em NeurIPS}, 2024.

\bibitem{llava15}
Haotian Liu, Chunyuan Li, Yuheng Li, and Yong~Jae Lee.
\newblock Improved baselines with visual instruction tuning.
\newblock In {\em CVPR}, 2024.

\bibitem{mmbench}
Yuan Liu, Haodong Duan, Yuanhan Zhang, Bo~Li, Songyang Zhang, Wangbo Zhao, Yike
  Yuan, Jiaqi Wang, Conghui He, Ziwei Liu, et~al.
\newblock Mmbench: Is your multi-modal model an all-around player?
\newblock In {\em ECCV}, 2024.

\bibitem{kivi2024}
Zirui Liu, Jiayi Yuan, Hongye Jin, Shaochen Zhong, Zhaozhuo Xu, Vladimir
  Braverman, Beidi Chen, and Xia Hu.
\newblock {KIVI}: A tuning-free asymmetric 2bit quantization for {KV} cache.
\newblock In {\em Proceedings of the 41st International Conference on Machine
  Learning}, volume 235 of {\em Proceedings of Machine Learning Research},
  pages 32332--32344. PMLR, 2024.

\bibitem{elasticcache2024}
Zuyan Liu, Benlin Liu, Jiahui Wang, Yuhao Dong, Guangyi Chen, Yongming Rao,
  Ranjay Krishna, and Jiwen Lu.
\newblock Efficient inference of vision instruction-following models with
  elastic cache.
\newblock In {\em ECCV}, 2024.
\newblock arXiv:2407.18121.

\bibitem{scienceqa}
Pan Lu, Swaroop Mishra, Tanglin Xia, Liang Qiu, Kai-Wei Chang, Song-Chun Zhu,
  Oyvind Tafjord, Peter Clark, and Ashwin Kalyan.
\newblock Learn to explain: Multimodal reasoning via thought chains for science
  question answering.
\newblock {\em NeurIPS}, 2022.

\bibitem{mirzasoleiman2015}
Baharan Mirzasoleiman, Ashwinkumar Badanidiyuru, Amin Karbasi, Jan Vondr{\'a}k,
  and Andreas Krause.
\newblock Lazier than lazy greedy.
\newblock In {\em Proceedings of the AAAI Conference on Artificial
  Intelligence}, 2015.
\newblock arXiv:1409.7938.

\bibitem{hubness2010}
Milo{\v s} Radovanovi{\'c}, Alexandros Nanopoulos, and Mirjana Ivanovi{\'c}.
\newblock Hubs in space: Popular nearest neighbors in high-dimensional data.
\newblock {\em Journal of Machine Learning Research}, 11:2487--2531, 2010.

\bibitem{renyi1961}
Alfr{\'e}d R{\'e}nyi.
\newblock On measures of entropy and information.
\newblock {\em Proc.\ 4th Berkeley Symp.\ on Mathematical Statistics and
  Probability}, 1961.

\bibitem{senerSavarese2018}
Ozan Sener and Silvio Savarese.
\newblock Active learning for convolutional neural networks: A core-set
  approach.
\newblock In {\em International Conference on Learning Representations (ICLR)},
  2018.
\newblock arXiv:1708.00489.

\bibitem{prumerge}
Yuzhang Shang, Mu~Cai, Bingxin Xu, Yong~Jae Lee, and Yan Yan.
\newblock {LLaVA-PruMerge}: Adaptive token reduction for efficient large
  multimodal models.
\newblock In {\em ICCV}, 2025.
\newblock arXiv:2403.15388.

\bibitem{textvqa}
Amanpreet Singh, Vivek Natarajan, Meet Shah, Yu~Jiang, Xinlei Chen, Dhruv
  Batra, Devi Parikh, and Marcus Rohrbach.
\newblock Towards vqa models that can read.
\newblock In {\em CVPR}, 2019.

\bibitem{quest2024}
Jiaming Tang, Yilong Zhao, Kan Zhu, Guangxuan Xiao, Baris Kasikci, and Song
  Han.
\newblock {QUEST}: Query-aware sparsity for efficient long-context {LLM}
  inference.
\newblock In {\em Proceedings of the 41st International Conference on Machine
  Learning}, volume 235 of {\em Proceedings of Machine Learning Research},
  pages 47901--47911. PMLR, 2024.

\bibitem{lookm2024}
Zhongwei Wan, Ziang Wu, Che Liu, Jinfa Huang, Zhihong Zhu, Peng Jin, Longyue
  Wang, and Li~Yuan.
\newblock {LOOK}-{M}: Look-once optimization in {KV} cache for efficient
  multimodal long-context inference.
\newblock In {\em Findings of the Association for Computational Linguistics:
  EMNLP 2024}, pages 4065--4078, Miami, Florida, USA, November 2024.
  Association for Computational Linguistics.

\bibitem{internvl35}
Weiyun Wang, Zhangwei Gao, Lixin Gu, Hengjun Pu, Long Cui, Xingguang Wei,
  Zhaoyang Liu, Linglin Jing, Shenglong Ye, Jie Shao, et~al.
\newblock Internvl3. 5: Advancing open-source multimodal models in versatility,
  reasoning, and efficiency.
\newblock {\em arXiv preprint arXiv:2508.18265}, 2025.

\bibitem{dart2025}
Zichen Wen, Yifeng Gao, Shaobo Wang, Junyuan Zhang, Qintong Zhang, Weijia Li,
  Conghui He, and Linfeng Zhang.
\newblock Stop looking for ``important tokens'' in multimodal language models:
  Duplication matters more.
\newblock In {\em Proceedings of the 2025 Conference on Empirical Methods in
  Natural Language Processing}, pages 9961--9980, Suzhou, China, November 2025.
  Association for Computational Linguistics.

\bibitem{streamingllm}
Guangxuan Xiao, Yuandong Tian, Beidi Chen, Song Han, and Mike Lewis.
\newblock Efficient streaming language models with attention sinks.
\newblock In {\em ICLR}, 2024.

\bibitem{visionzip2025}
Senqiao Yang, Yukang Chen, Zhuotao Tian, Chengyao Wang, Jingyao Li, Bei Yu, and
  Jiaya Jia.
\newblock {VisionZip}: Longer is better but not necessary in vision language
  models.
\newblock In {\em CVPR}, 2025.
\newblock arXiv:2412.04467.

\bibitem{fitprune}
Weihao Ye, Qiong Wu, Wenhao Lin, and Yiyi Zhou.
\newblock Fit and prune: Fast and training-free visual token pruning for
  multi-modal large language models.
\newblock In {\em AAAI}, 2025.
\newblock arXiv:2409.10197.

\bibitem{mmvet}
Weihao Yu, Zhengyuan Yang, Linjie Li, Jianfeng Wang, Kevin Lin, Zicheng Liu,
  Xinchao Wang, and Lijuan Wang.
\newblock {MM-Vet}: Evaluating large multimodal models for integrated
  capabilities.
\newblock In {\em ICML}, 2024.

\bibitem{h2o2023}
Zhenyu Zhang, Ying Sheng, Tianyi Zhou, Tianlong Chen, Lianmin Zheng, Ruisi Cai,
  Zhao Song, Yuandong Tian, Christopher R{\'e}, Clark Barrett, et~al.
\newblock H2o: Heavy-hitter oracle for efficient generative inference of large
  language models.
\newblock {\em NeurIPS}, 2023.

\end{thebibliography}
\end{document}